# Improving Bias in Facial Attribute Classification: A Combined Impact of KL Divergence induced Loss Function and Dual Attention

Shweta Patel[1]  and Dakshina Ranjan Kisku[2]

Department of Computer Science and Engineering
National Institute of Technology Durgapur[1,2]
Durgapur-713209, India
sp.21cs1110@phd.nitdgp.ac.in
drkisku.cse@nitdgp.ac.in

**Abstract.** Ensuring that AI-based facial recognition systems produce fair predictions and work equally well across all demographic groups is crucial. Earlier systems often exhibited demographic bias, particularly in gender and racial classification, with lower accuracy for women and individuals with darker skin tones. To tackle this issue and promote fairness in facial recognition, researchers have introduced several bias-mitigation techniques for gender classification and related algorithms. However, many challenges remain, such as data diversity, balancing fairness with accuracy, disparity, and bias measurement. This paper presents a method using a dual attention mechanism with a pre-trained Inception-ResNet V1 model, enhanced by KL-divergence regularization and a cross-entropy loss function. This approach reduces bias while improving accuracy and computational efficiency through transfer learning. The experimental results show significant improvements in both fairness and classification accuracy, providing promising advances in addressing bias and enhancing the reliability of facial recognition systems.

**Keywords:** Fairness · Face recognition · Demographic bias · KL-Divergence · Fine-tuning · CNN

## 1 Introduction

Facial recognition(FR) technology [5] is commonly used for both authentication and identification purposes, but performance inconsistencies across different demographic groups raise concerns about fairness. These biases disproportionately affect underrepresented populations, potentially restricting their access to important services and resources. As AI systems are increasingly applied in critical sectors like mortgage approval and criminal justice, preventing discriminatory outcomes becomes crucial [4]. A survey by NIST's FRVT [1] highlighted substantial performance differences based on gender and race, underscoring ethical issues. Addressing these biases is crucial to ensure FR systems are fair and reliable, prompting ongoing research to reduce such disparities. Demographic bias in FR systems often arises from multiple sources, with unbalanced datasets [18] being a major contributor. Other contributing factors include inadequate data processing, flawed model structures, and improper application of machine learning (ML) or deep learning (DL) techniques. Research frequently points to datasets that overrepresent specific groups, such as lighter-skinned males, while underrepresenting darker-skinned females. Even with balanced datasets, some techniques still exhibit skewed performance, particularly for black individuals [22]. To address these problems, various post-processing methods have been suggested, such as score normalization across demographic groups [2] and leveraging facial attributes as supplementary data to enhance the model's probability predictions and overall accuracy  [3].

The proposed model presents a new method by incorporating score distribution probability and face embedding as a regularization term, paired with an attention mechanism to reduce bias and improve accuracy. This regularization term reduces the gap between predicted and actual probabilities, ensuring a normalized score distribution and fairer decisions, while also correcting prediction errors. As far as we know, this is the first approach to integrate distribution scores into the objective function. The attention mechanism helps focus more effectively on underrepresented groups, and the regularization terms further mitigate bias. By combining data handling techniques with architectural modifications in a pre-trained model like Inception-ResNet v1, our approach shows significant improvements in both fairness metrics and classification accuracy over existing methods. These findings indicate that our model can contribute to the development of fairer facial attribute recognition systems, addressing long-standing biases in the field.

The significant findings of the proposed work are:

- The proposed method presents a novel dual attention mechanism integrated into the Inception-ResNet V1 model. This mechanism improves feature extraction by focusing on the most critical spatial areas and feature channels, leading to enhanced accuracy and reduced bias.
- A Kullback-Leibler (KL) divergence induced regularization term is added to the cross-entropy loss function. This method aids in generating more calibrated and less biased predictions by aligning the predicted probability distribution with the intended target probability distribution.
- Another novel addition as a regularization term is introduced to minimize the intra-class distance between embeddings from different demographic groups. This reduces bias by encouraging tighter clustering of samples within the same demographic group.
- The proposed work introduces a new bias measurement analysis plot that offers insights into how bias is reduced across various demographic groups, especially in terms of race and gender classifications.
- The work highlights the importance of using balanced datasets, such as Fair Face, UTK Face, and BFW, for training and evaluation to prevent the model from inheriting biases caused by imbalanced data. These actions together strive to develop a more equitable facial recognition system by tackling the persistent bias issues found in existing models.

The structure of this paper is as follows: Section 2 provides a detailed litera-ture review that has been studied in the related works section, which identifies the open challenges with possible solutions. Section 3 describes the proposed framework. Section 4 presents the experimental setup which includes datasets, pre-processing and evaluation metrics. Section 5 reports experimental results, analysis and discussion and comparison of the proposed framework with SOTA models. Finally, Section 6 concludes the paper and suggests directions for future research.

## 2    Related Works

This section offers a comprehensive review of some existing works, systemat-ically structured into two primary subsections: a) Classical Machine Learning approaches and b) Deep Learning approaches.

### 2.1    Classical Machine Learning techniques

Research in computer vision has long focused on identifying gender and demo-graphic attributes from facial images. The authors of [16] proposed a method for extracting primary and secondary facial features for race classification using the Viola-Jones method for face detection and Sobel edge operators for regions like the forehead, eyes, and lips. However, this method was limited by a small, constrained dataset. The method discussed in [12] utilized biologically inspired features with Gabor filters, but accuracy decreased when training on females and testing on males, likely due to dataset imbalance. In contrast, [13] extracted 77 facial landmarks from the "T" region of the face, using the mRMR algo-rithm and k-NN for demographic classification. While accurate, this method was limited to frontal face detection and performed poorly on side faces. The work reported in [14] used periorbital features from facial images and proposed an ethnicity recognition system, employing a level co-occurrence matrix, colour histogram for feature extraction, and a random forest classifier. However, this study used a private dataset and lacked state-of-the-art comparative analysis. A work [15] that introduced a compact-fusion features framework for ethnicity classification, utilizing four handcrafted techniques for feature selection, embed-ding extraction, and SVM one-vs-all classification. While effective, this method does not control the feature reduction ratio and struggles with large-scale data due to high computing demands. To address these issues, hybrid techniques have been developed. For example, a combination of ML-based SVM classifier with DL-based feature extractors [17], used a pre-trained VGG16 model. Effective hybrid models require a thorough understanding of both the problem and the involved models.

### 2.2    Deep Learning approaches

The advancement in deep learning methods like adversarial learning, variational autoencoder, transfer learning, etc., has led to significant achievements in miti-gating bias in face recognition. The work [8] assessed the performance of several CNN architectures in gender classification across gender-racial groupings. They have used various pre-trained models to accelerate the comparison between the architectures. The authors proposed that disparities in architecture had an impact on uneven accuracy rates. The high misclassification error rate of black females is thought to cause their significant physical similarity. The research [28] incorporates a joint loss function to recognise the ethnicity that combines softmax loss with the weighted centre loss function, which uses an attention learning module to select the most distinguished features in the embedding space. The joint loss study relies solely on overall accuracy as an evaluation metric, which fails to capture equivalent performance across different demographic groups. No metric in this study reflects equal performance across all ethnic groups. The authors [29] proposed a method to reduce bias in diverse demographic populations by aligning facial features at multiple levels for both healthy individuals and those with various disorders. Their approach aimed to enhance fairness by focusing on the localization of facial landmarks. However, this method is sensitive to local occlusions and struggles to perform well on unconstrained datasets. The authors of [9] proposed methods based on generative views, utilizing GAN-based latent vector editing combined with structured learning. A neural network pruning method [10] that calculates the per-group relevance of each model weight. The approach iteratively selects and prunes weights with lower relevance values to reduce performance discrepancies. The FairFace [18], UTKFace [20], and CelebA [7] datasets are used to demonstrate the effectiveness of this method. Additionally, Fair Supervised Contrastive Loss (FSCL) was introduced in [11], which ensures fair representation learning by enforcing that representations from the same class are closer to each other than those from other classes.

## 3   Proposed Methodology

The proposed work achieves unbiased and enhanced performance for the face attribute classification. Merely using a balanced dataset across different demographics will not give an unbiased performance; specific, explicit criteria must be introduced to deal with the algorithmic bias. In this section, we discuss the proposed approach that uses a customised loss function to mitigate the bias in race and gender classification by enhancing the critical features and suppressing the irrelevant features using an attention mechanism. Figure 1 illustrates the detailed flow of the model.

### 3.1   Overview

The aim of the proposed work is to train a model that minimizes statistically significant performance variation across different groups. Such variations can stem from image-related factors, like pose, illumination, and resolution, which introduce intrinsic bias, and from subject-related factors, like race, gender, and age, which lead to algorithmic bias affecting facial attribute recognition. Our focus is on reducing algorithmic bias inherent in the model itself. Performance variations can also arise from imbalanced datasets, which prevent the network from effectively learning salient features of underrepresented groups. While a balanced dataset across demographic attributes can help, it alone is not sufficient. We must also introduce specific criteria to ensure the model addresses and corrects biased performance.

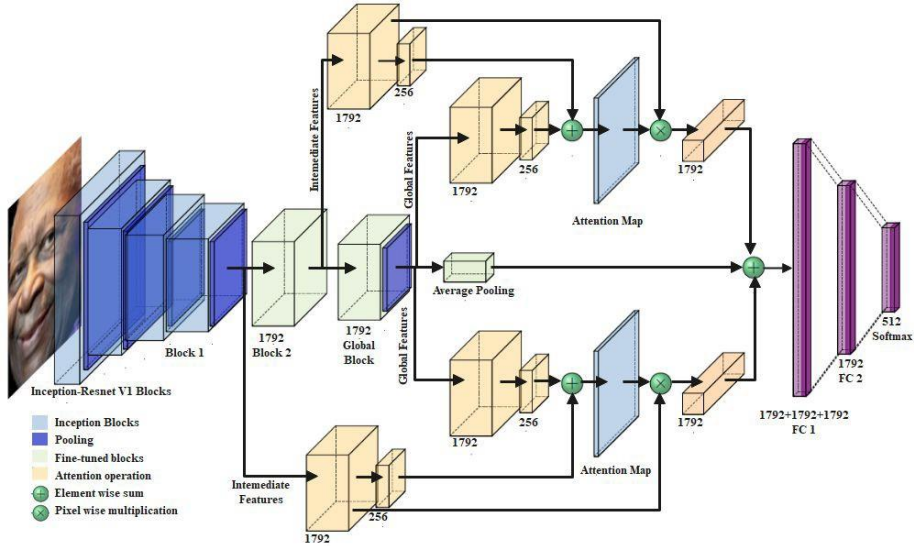

**Fig. 1:** Block diagram of the proposed model.

This study introduces and evaluates an attention-based face attribute classification model specifically designed to enhance accuracy and address bias through a custom loss function. The Squeeze-and-Excitation (SE) [31] attention mechanism captures relevant information by adaptive weighting and rescaling the channels, allowing the network to focus on the most informative features. The model incorporates trainable attention modules to extract features that the Inception-ResNet V1 [23] architecture, pre-trained on the VGGFace2 dataset [23], cannot capture effectively. This hybrid architecture combines Inception and Residual networks with skip connections to enhance feature extraction across various scales. The approach integrates attention with multiple loss functions to minimize classification loss, reduce intra-class embedding distance, and align predicted with target probability distributions, thereby improving performance and fairness. The model emphasizes increasing confidence scores to promote equity and reduce bias.

### 3.2   Attention Characterized Map Generation

The human vision focuses on features related to the task at hand in its purview. If we want to recognise the gender of the person, we will focus on the features that are relevant to gender, like hair, jawline, and the texture of the skin and for ethnicity, we focus on the T-area (including the distance between eyebrows, nose, and lips shape), skin colour etc. Likewise, we are using attention modules to estimate feature maps to make the model emphasise the relevant areas. To focus on the race-related features, we target feature channels by using channel-wise attention. Channel-wise attention is designed to focus on the most important channel. After getting the channel-focused features, we passed them through the spatial attention with an element-wise additive fusion mechanism, where local and global features are combined additively before applying spacial attention.

### 3.3   KL-Divergence Induced De-biasing Loss Function

After feature extraction, the network was optimized on three criteria: minimizing cross-entropy loss between predicted and target logits, reducing the variation in confidence scores of predictions, and decreasing the intra-class distance between positive embeddings. A KL divergence-regularized term was added to the cross-entropy loss to account for all probability values. KL divergence [32], a measure of how one probability distribution differs from a reference, is non-negative and reaches zero only when distributions are identical. While cross-entropy loss typically focuses on the highest probability, incorporating KL divergence adds flexibility, improving model calibration and generalization, particularly when labels are uncertain or regularization is applied. This approach helps produce more accurate and less biased predictions by aligning the predicted distribution with a target distribution. The Classification Loss is defined as follows:

$$L_{\text{combined}} = -\sum_{i=1}^{N} y_i \log(\hat{y}_i) + \lambda \sum_{i=1}^{N} P(i) \log \frac{P(i)}{Q(i)} \tag{1}$$

where $N$ denotes the number of samples, $y_i$ denotes the true label for the $i$-th sample, $\hat{y}_i$ denotes the predicted probability for the $i$-th sample, $\lambda$ denotes the regularised term for KL-Divergence loss, $P(i)$ denotes the actual confidence score for the $i$-th sample and $Q(i)$ denotes the predicted confidence score for the $i$-th sample.

In addition to the face attribute classification objective function, a regularization term is included to reduce the intra-class distance between demographic groups. The Mahalanobis distance is used to measure intra-class distance for each race. To achieve this, the following loss components are defined:

$$L_{\text{intra-class}} = \beta \frac{1}{N} \sum_{c \in C} \sum_{i \in I_c} \sqrt{(e_i - \mu_c)^T S^{-1}(e_i - \mu_c)} \tag{2}$$

where C represents set of all classes, $I_c$ denotes set of indices of samples in class c, $e_i$ is the embedding of sample i, $\mu_c$ denotes mean embedding (centre) of class c, S denotes covariance matrix of all embeddings, $S^{-1}$ represents the inverse of the covariance matrix, $N$ is the total number of intra-class pairs and $\beta$ denotes hyperparameter.

### 3.4  Working principle

The attention mechanism is primarily concerned with identifying the most critical aspects of the input data. This approach is also useful for distinguishing between different classes by recognizing their unique characteristics. KL divergence, on the other hand, quantifies the disparity between the target probability distribution and the predicted probability distribution, motivating the model to generate a probability distribution that is more similar to the intended distribution. The intra-class compactness promotes the clustering of embeddings that belong to the same class, thereby reducing variability within the same class and enhancing the discriminative features. In conjunction with these regularization terms, the attention mechanism enables the model to focus on the most salient features, enhance accuracy by considering the probability distribution, and narrow the gap between the intra-class distance of different classes. We reduce the number of incorrect predictions by utilizing the KL-regularization term.

## 4  Experimental Setup

### 4.1  Dataset used

The proposed model is trained using the gender- and race-balanced FairFace dataset to reduce classification bias. To evaluate the model, we use the balanced FairFace [18], UTKFace [20], and BFW [21] datasets. These datasets ensure that evaluations are unbiased and equally distributed across genders and races. They include diverse photos varying in age, gender, pose, lighting, and expression. A detailed description of the datasets is provided below:

**Table 1:** Distribution of Training set among different races considered for FairFace Dataset.

| Race | Male | Female | Total |
|------|------|--------|-------|
| White | 8701 | 7826 | 16527 |
| Black | 6096 | 6137 | 12233 |
| East Asian | 6146 | 6141 | 12287 |
| Indian | 6410 | 5909 | 12319 |
| Total | 27353 | 26009 | 53362 |

**Table 2:** Distribution of Testing set among different races considered for FairFace Dataset.

| Race | Male | Female | Total |
|------|------|--------|-------|
| White | 1122 | 963 | 2085 |
| Black | 799 | 757 | 1556 |
| East Asian | 777 | 773 | 1550 |
| Indian | 753 | 763 | 1516 |
| Total | 3451 | 3256 | 6707 |

**FairFace:** The FairFace dataset contains 108,501 photos with a balanced composition across race and gender. It includes seven ethnic groups—White, Black, Indian, East Asian, Southeast Asian, Middle Eastern, and Latino Hispanic—covering both genders and various ages. The dataset comprises 53% men and 47% women. For this experiment, images of White, Black, Indian, and East Asian races were used. The training set distribution is detailed in Table 1, the details of testing set is given in Table 2, and sample photos are shown in Figure 2.

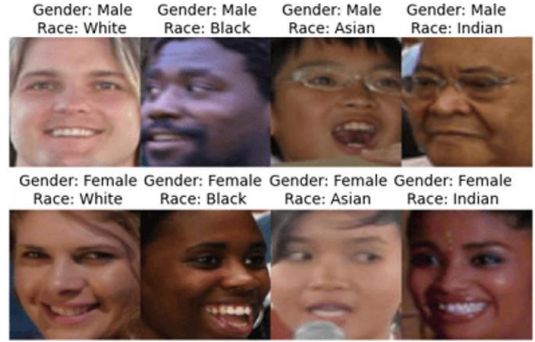

**Fig. 2:** Some sample faces cropped by MTCNN from Faiface Dataset

**UTKFace:** The UTKFace dataset includes over 20,000 face images with annotations for age, gender, and ethnicity (White, Black, Asian, Indian, and Others). Due to ambiguity, the "Other" category was excluded, and some White ethnicity images were removed to balance the dataset for evaluation. The image distribution by race and gender is shown in Table 3, with samples in Figure 3.

**Table 3:** Distribution of Testing set among different races considered for UTKFace Dataset.

| Race | Male | Female | Total |
|------|------|--------|-------|
| White | 2605 | 2390 | 4995 |
| Black | 2338 | 2223 | 4561 |
| Asian | 1644 | 1942 | 3586 |
| Indian | 2285 | 1742 | 4027 |
| Total | 8872 | 8297 | 17169 |

**Table 4:** Distribution of Testing set among different races considered for BFW Dataset.

| Race | Male | Female | Total |
|------|------|--------|-------|
| White | 2500 | 2500 | 5000 |
| Black | 2500 | 2500 | 5000 |
| Asian | 2500 | 2500 | 5000 |
| Indian | 2500 | 2500 | 5000 |
| Total | 10000 | 10000 | 20000 |

**BFW dataset** The BFW dataset was used for additional proposed model evaluation, featuring balanced representation by race and gender. It includes 5,000 faces per race across Asian, African, Caucasian, and Indian categories, with a 50% female and 50% male distribution for each race. Image distribution details are in Table 4, and sample images are shown in Figure 4.

### 4.2   Pre-processing

All face images are cropped and resized to 240 × 240 pixels using Multi-Task Cascaded Convolutional Neural Networks (MTCNN) [24]. The dataset contains images from uncontrolled environments, making face detection challenging due to variations in size, pose, illumination, and background. To improve face extraction, we adjusted MTCNN thresholds and used Contrast Limited Adaptive Histogram Equalization (CLAHE) [25] for local contrast enhancement. To reduce overfitting and ensure fairness across gender and racial groups, we applied facial data transformations such as random horizontal flipping and affine transformations. Balanced batch [27] training further prevents bias by ensuring equal influence of each class on model updates, promoting better generalization and fairness.

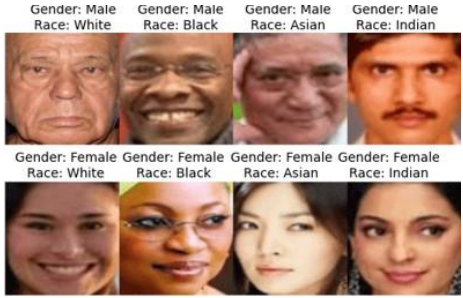

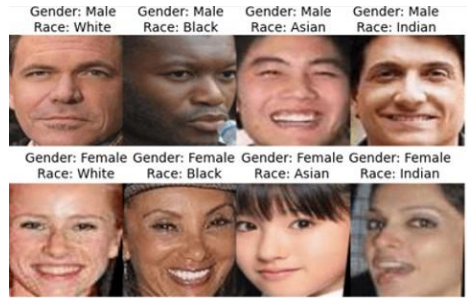

**Fig. 3:** Some sample face images cropped by MTCNN from UTKFace Dataset

**Fig. 4:** Some sample face images cropped by MTCNN from BFW Dataset

### 4.3    Implementation details

The proposed model is trained on the FairFace dataset using a pre-trained Inception-ResNet V1 model, initially trained on the VGGFace2 [26] dataset, to extract key facial features. For evaluation, we use the UTKFace and BFW datasets, which are balanced for race and gender. These datasets are pre-processed in the same manner as the FairFace dataset. Training is optimized with the AdamW optimizer, a weight decay of 0.002, and a batch size of 64. The learning rate starts at 0.0001 and decays to 1e-7 with patience set to 7 and a factor of 0.1, reducing the learning rate if the validation loss does not improve for seven consecutive epochs. The model is trained to produce a 512-dimensional embedding representation. On an Nvidia Tesla V100 GPU, the average training speed is 6.6 images per second, with a total of 7.5 million model parameters.

### 4.4    Evaluation Metrics

We evaluated the model's performance using classwise classification accuracy and overall accuracy, comparing our results to State-Of-The-Art (SOTA) methods. Additionally, we employed the Degree of Bias (DoB), which calculates the standard deviation of classification accuracy across facial attribute subgroups. Lower DoB values indicate less bias, while higher values reflect greater bias, signifying accuracy variation between target groups. To further assess uneven performance, we used the Max/Min ratio, defined as the ratio of maximum to minimum accuracy across race groups, where a larger ratio indicates greater disparity.

## 5   Evaluation

### 5.1   Experimental Results

The proposed attention-based hybrid model was evaluated on the FairFace test subset for bias estimation across ethnicity-gender groups, followed by testing on UTKFace and BFW datasets. Figure 5 shows the training accuracy curve, with initial fluctuations due to regularization techniques like dropout and data augmentation to prevent overfitting. The model achieved training accuracies of 97.45% for gender classification and 95.12% for race classification, with validation accuracies of 97.28% and 95.07%, respectively. Performance was measured using metrics sensitive to both prediction accuracy and confidence.

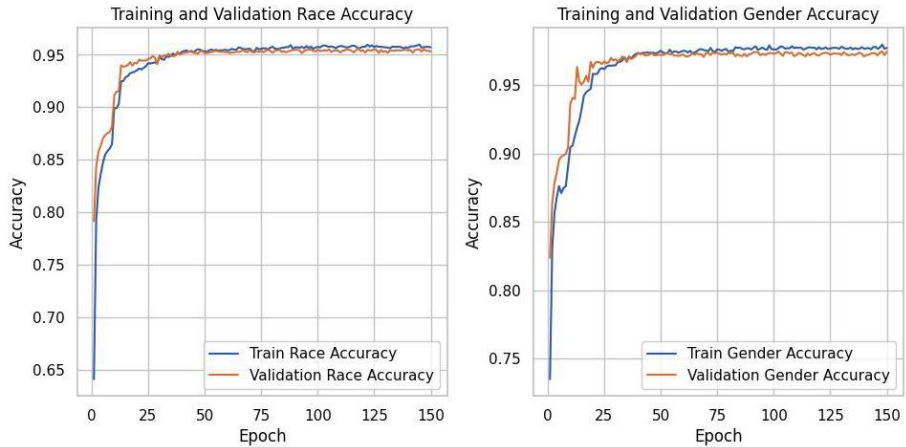

**Fig. 5:** Accuracy curve

### 5.2   Analysis and Discussion

The attention mechanism in our model effectively extracts and amplifies complex patterns, allowing it to match or exceed the performance of state-of-the-art (SOTA) methods. Figure 6 shows the raw augmented input images, features highlighted by the attention module, which enhances features from each channel of the intermediate feature map with higher values. The ReLU activation function in the attention network ensures that only prominent features are retained.

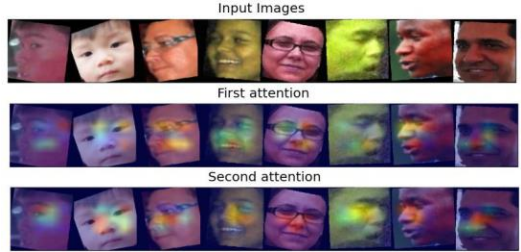

**Fig. 6:** Features extracted by both attention mechanism

The regularization techniques have effectively addressed the generalization-fairness trade-off. The use of class-balanced batches mitigates bias at the batch level, and additionally, the use of the intra-class loss further enhances model fairness and consistency.

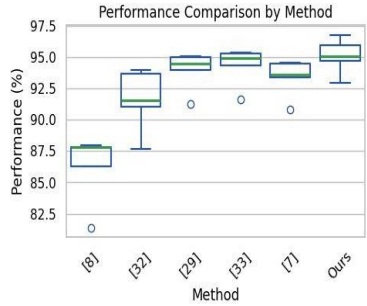

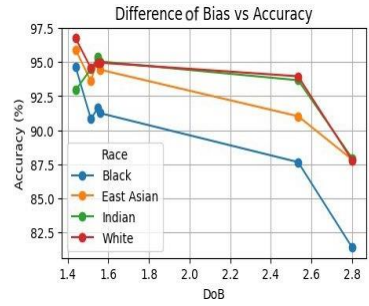

**Fig. 7:** Performance of Different Methods Across Different Races

**Fig. 8:** Difference of Bias(DoB) vs Accuracy

Figure 7 displays the accuracy distribution of state-of-the-art (SOTA) methods across various racial groups. The proposed approach demonstrates accuracies clustered into small regions without outliers, whereas other methods show a wider dispersion of results. This indicates that the proposed work achieves a better balance between generalization and consistent performance across all racial groups.

To illustrate the problem further, a new plot is introduced showing the degree of bias relative to the accuracy across various racial groups, quantifying accuracy disparities. Figure 8 depicts how racial accuracy differences affect coverage of differences of bias (DoB). The difference of bias decreases as racial accuracies become more similar and increases when there are significant differences. Despite some races, such as White, East Asian, and Indian, showing closer accuracy in the SOTA method, the Black race remains less aligned with others, resulting in higher bias. This discrepancy highlights that the methodologies do not perform equally well across all races. Whereas for the proposed work, accuracies of all races are in closed-distance resulting in lower degree of bias.

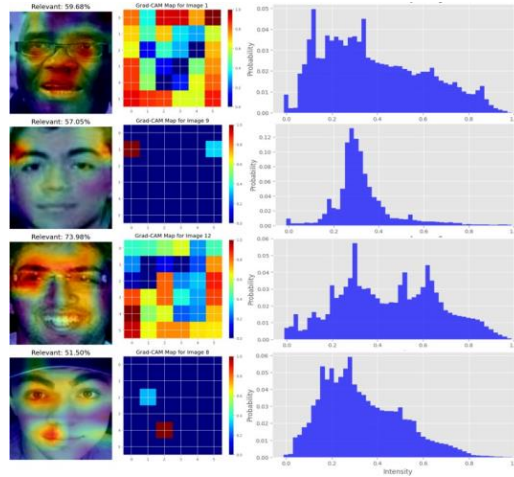

**Fig. 9:** Relavancy depicted using GRAD-CAM method

The Grad-CAM [36] method is used to visualize the regions responsible for race prediction. Figure 9 shows the spread of relevant features used by the model. Grad-CAM generates a heatmap by using the gradients of the target class flowing into the last convolutional layer. The feature maps are weighted by these gradients, with higher intensity in the heatmap indicating areas with a greater impact on the final prediction.

### 5.3    Comparison with SOTA models

Table 5 shows the comparative analysis of the proposed model with some SOTA methods for overall classification accuracy for race attribute, the Degree of bias and the max/min ratio, which depicts disparity as metrics. The table shows that the proposed approach comprises a dual attention mechanism, regularized loss function and the class balanced batch that has achieved remarkable performance compared with other SOTA methods. The proposed model has obtained the highest overall classification accuracy and least DoB which represents the model is able to mitigate the bias in race classification. Though the max/min ratio is not the least one, but comparable with other methods. Although the results for the BFW dataset are lower compared to the Fairface and UTKFace datasets, this is due to the BFW dataset having fewer samples and less diversity in terms of poses and illumination, which limits the model's ability to generalize and perform well.

**Table 5:** Comparative analysis with SOTA approaches for race classification

| Method | Black | East Asian | Indian | White | Overall | DoB | Max/Min |
|---|---|---|---|---|---|---|---|
| FairFace | | | | | | | |
| Pre-trained [8] | 81.4 | 87.85 | 87.95 | 87.8 | 86.25 | 2.8 | 1.08 |
| Adversarial debiasing [33] | 87.66 | 91.93 | 93.67 | 93.96 | 91.80 | 2.51 | 1.071 |
| Multi-Tasking [30] | 91.26 | 94.45 | 95.05 | 94.96 | 93.93 | 1.55 | 1.041 |
| Deep Generative Views based [34] | 91.64 | 95.29 | **95.38** | 94.92 | 94.30 | 1.54 | **1.040** |
| Regularization-based technique [6] | 90.83 | 93.6 | 94.48 | 94.57 | 93.37 | 1.51 | 1.041 |
| Proposed method | **94.68** | **95.93** | 92.94 | **96.76** | **95.07** | **1.43** | 1.041 |
| UTKFace | | | | | | | |
| Adversarial debiasing [33] | **94.62** | - | 93.65 | 94.97 | 93.78 | 1.38 | 1.03 |
| regularization-based technique [6] | 95.85 | - | **95.43** | 95.16 | 95.03 | **0.95** | **1.02** |
| Proposed method | 92.26 | **94.45** | 95.05 | **95.96** | **96.64** | 1.81 | 1.067 |
| BFW | | | | | | | |
| Proposed method | 95.25 | 94.85 | 93.59 | 96.08 | 94.94 | 0.89 | 1.026 |

## 6   Conclusion and Future Scope

To mitigate the bias in facial attribute recognition, we have introduced a novel method comprised of an attention mechanism and custom loss function that minimises the losses at logits, confidence score and embedding level. The experimental evaluations demonstrate that the proposed model significantly enhances both fairness measures and classification accuracy compared to existing methods. Specifically, the attention mechanism reduces bias effectively, and the custom loss function improves overall performance, facilitating equitable facial recognition. Pre-trained models, often biased due to their training datasets, can affect subsequent tasks and may be difficult to interpret due to their "black box" nature. In the future, we plan to train the model from scratch on a balanced and diverse dataset. Additionally, most available datasets focus on binary gender groups, but developing a fair system requires considering all gender identities, including LGBTQ+. Future research will address this underexplored area with some novel frameworks.

## References

1. Patrick Grother, Mei Ngan, and Kayee Hanaoka. Face recognition vendor test (fvrt): Part 3, demographic effects. National Institute of Standards and Technology Gaithersburg, MD, 2019.
2. Jain, A., Nandakumar, K., & Ross, A. (2005). Score normalization in multimodal biometric systems. Pattern recognition, 38(12), 2270-2285.
3. Yang, Z., Zhu, X., Jiang, C., Liu, W., & Shen, L. (2021, August). Ramface: Race adaptive margin based face recognition for racial bias mitigation. In 2021 IEEE International Joint Conference on Biometrics (IJCB) (pp. 1-8). IEEE.
4. Ahmed, M. A., Choudhury, R. D., & Kashyap, K. (2022). Race estimation with deep networks. Journal of King Saud University-Computer and Information Sciences, 34(7), 4579-4591.

5. Cavazos, J. G., Phillips, P. J., Castillo, C. D., & O'Toole, A. J. (2020). Accuracy comparison across face recognition algorithms: Where are we on measuring race bias?. IEEE transactions on biometrics, behavior, and identity science, 3(1), 101-111.

6. Krishnan, A., & Rattani, A. (2023). A novel approach for bias mitigation of gender classification algorithms using consistency regularization. Image and Vision Computing, 137, 104793.

7. Liu, Z., Luo, P., Wang, X., & Tang, X. (2015). Deep learning face attributes in the wild. In Proceedings of the IEEE international conference on computer vision (pp. 3730-3738).

8. Krishnan, A., Almadan, A., & Rattani, A. (2020, December). Understanding fairness of gender classification algorithms across gender-race groups. In 2020 19th IEEE international conference on machine learning and applications (ICMLA) (pp. 1028-1035). IEEE.

9. Ramachandran, S., & Rattani, A. (2022, August). Deep generative views to mitigate gender classification bias across gender-race groups. In International Conference on Pattern Recognition (pp. 551-569). Cham: Springer Nature Switzerland.

10. Lin, X., Kim, S., & Joo, J. (2022, October). Fairgrape: Fairness-aware gradient pruning method for face attribute classification. In European Conference on Computer Vision (pp. 414-432). Cham: Springer Nature Switzerland.

11. Park, S., Lee, J., Lee, P., Hwang, S., Kim, D., & Byun, H. (2022). Fair contrastive learning for facial attribute classification. In Proceedings of the IEEE/CVF Conference on Computer Vision and Pattern Recognition (pp. 10389-10398).

12. Guo, G., & Mu, G. (2010, June). A study of large-scale ethnicity estimation with gender and age variations. In 2010 IEEE Computer Society Conference on Computer Vision and Pattern Recognition-Workshops (pp. 79-86). IEEE.

13. Wang, C., Zhang, Q., Liu, W., Liu, Y., & Miao, L. (2019). Expression of Concern: Facial feature discovery for ethnicity recognition. Wiley Interdisciplinary Reviews: Data Mining and Knowledge Discovery, 9(1), e1278.

14. Putriany, D. M., Rachmawati, E., & Sthevanie, F. (2021, February). Indonesian ethnicity recognition based on face image using gray level co-occurrence matrix and color histogram. In IOP conference series: materials science and engineering (Vol. 1077, No. 1, p. 012040). IOP Publishing.

15. Wirayuda, T. A. B., Munir, R., & Kistijantoro, A. I. (2023, June). Compact-Fusion Feature Framework for Ethnicity Classification. In Informatics (Vol. 10, No. 2, p. 51). MDPI.

16. Roomi, S. M. M., Virasundarii, S. L., Selvamegala, S., Jeevanandham, S., & Hariharasudhan, D. (2011, December). Race classification based on facial features. In 2011 third national conference on computer vision, pattern recognition, image processing and graphics (pp. 54-57). IEEE.

17. Ache, A. I., Kassir, M. M., & Ibrahim, H. M. (2023). Ethnicity Classification: Discriminating Between Iranian and Asian Populations Using Hybrid Deep Learning Algorithm.

18. Karkkainen, K., & Joo, J. (2021). Fairface Face attribute dataset for balanced race, gender, and age for bias measurement and mitigation. In Proceedings of the IEEE/CVF winter conference on applications of computer vision (pp. 1548-1558).

19. Vera-Rodriguez, R., & Tolosana, R. (2021). SensitiveNets: Learning Agnostic Representations with Application to Face Images. IEEE TRANSACTIONS ON PATTERN ANALYSIS AND MACHINE INTELLIGENCE, 43(6).

20. Zhang, Z., Song, Y., & Qi, H. (2017). Age progression/regression by conditional adversarial autoencoder. In Proceedings of the IEEE conference on computer vision and pattern recognition (pp. 5810-5818).

21. Robinson, J. P., Qin, C., Henon, Y., Timoner, S., & Fu, Y. (2023). + IEEE Transactions on Image Processing.

22. Wang, M., Zhang, Y., & Deng, W. (2021). Meta balanced network for fair face recognition. IEEE transactions on pattern analysis and machine intelligence, 44(11), 8433-8448.

23. Szegedy, C., Ioffe, S., Vanhoucke, V., & Alemi, A. (2017, February). Inception-v4, inception-resnet and the impact of residual connections on learning. In Proceedings of the AAAI conference on artificial intelligence (Vol. 31, No. 1).

24. Zhang, K., Zhang, Z., Li, Z., & Qiao, Y. (2016). Joint face detection and alignment using multitask cascaded convolutional networks. IEEE signal processing letters, 23(10), 1499-1503.

25. Zuiderveld, K. (1994). Contrast limited adaptive histogram equalization. In Graphics gems IV (pp. 474-485).

26. Cao, Q., Shen, L., Xie, W., Parkhi, O. M., & Zisserman, A. (2018, May). Vggface2: A dataset for recognising faces across pose and age. In 2018 13th IEEE international conference on automatic face & gesture recognition (FG 2018) (pp. 67-74). IEEE.

27. Buda, M., Maki, A., & Mazurowski, M. A. (2018). A systematic study of the class imbalance problem in convolutional neural networks. Neural networks, 106, 249-259.

28. Dammak, S., Mliki, H., & Fendri, E. (2024, April). Facial Ethnicity Recognition Based on a New Joint Loss Function. In Asian Conference on Intelligent Information and Database Systems (pp. 169-180). Singapore: Springer Nature Singapore.

29. Freitas, R. T., Aires, K. R., de Paiva, A. C., Veras, R. D. M., & Soares, P. L. (2024). A CNN-based multi-level face alignment approach for mitigating demographic bias in clinical populations. Computational Statistics, 39(5), 2557-2579.

30. Das, A., Dantcheva, A., & Bremond, F. (2018). Mitigating bias in gender, age and ethnicity classification: a multi-task convolution neural network approach. In Proceedings of the european conference on computer vision (eccv) workshops (pp. 0-0).

31. Hu, J., Shen, L., & Sun, G. (2018). Squeeze-and-excitation networks. In Proceedings of the IEEE conference on computer vision and pattern recognition (pp. 7132-7141).

32. Csiszár, I. (1975). I-divergence geometry of probability distributions and minimization problems. The annals of probability, 146-158.

33. Zhang, B. H., Lemoine, B., & Mitchell, M. (2018, December). Mitigating unwanted biases with adversarial learning. In Proceedings of the 2018 AAAI/ACM Conference on AI, Ethics, and Society (pp. 335-340).

34. Ramachandran, S., & Rattani, A. (2022, August). Deep generative views to mitigate gender classification bias across gender-race groups. In International Conference on Pattern Recognition (pp. 551-569). Cham: Springer Nature Switzerland.

35. Das, A., Dantcheva, A., & Bremond, F. (2018). Mitigating bias in gender, age and ethnicity classification: a multi-task convolution neural network approach. In Proceedings of the european conference on computer vision (eccv) workshops (pp. 0-0).

36. Selvaraju, R. R., Cogswell, M., Das, A., Vedantam, R., Parikh, D., & Batra, D. (2020). Grad-CAM: visual explanations from deep networks via gradient-based localization. International journal of computer vision, 128, 336-359.